\documentclass{article}
\usepackage{spconf,amsmath,graphicx}

\usepackage{import}
\usepackage{booktabs}
\usepackage{tablefootnote}
\usepackage{amsfonts}
\usepackage{amssymb}
\usepackage{caption} 

\title{A Practical Two-Stage Training Strategy for multi-stream \\
end-to-end speech recognition}
\name{Ruizhi Li$^1$, Gregory Sell$^{1,2}$, Xiaofei Wang$^{3*}$\thanks{* This work was done while Xiaofei Wang was a post-doc at Johns Hopkins University.}, Shinji Watanabe$^1$, Hynek Hermansky$^{1,2}$}
\address{
  $^1$Center for Language and Speech Processing, The Johns Hopkins University, USA\\
  $^2$Human Language Technology Center of Excellence, The Johns Hopkins University, USA\\
  $^3$Speech and Dialog Research Group, Microsoft, USA}

\begin{document}
\ninept
\maketitle
\begin{abstract}

The multi-stream paradigm of audio processing, in which several sources are simultaneously considered, has been an active research area for information fusion.  
Our previous study offered a promising direction within end-to-end automatic speech recognition, where parallel encoders aim to capture diverse information followed by a stream-level fusion based on attention mechanisms to combine the different views.  
However, with an increasing number of streams resulting in an increasing number of encoders, the previous approach could require substantial memory and massive amounts of parallel data for joint training. 
In this work, we propose a practical two-stage training scheme.
Stage-1 is to train a Universal Feature Extractor (UFE), where encoder outputs are produced from a single-stream model trained with all data. 
Stage-2 formulates a multi-stream scheme intending to solely train the attention fusion module using the UFE features and pretrained components from Stage-1.
Experiments have been conducted on two datasets, DIRHA and AMI, as a multi-stream scenario.
Compared with our previous method, this strategy achieves relative word error rate reductions of 8.2--32.4\%, while consistently outperforming several conventional combination methods. 

\end{abstract}
\begin{keywords}
End-to-End Speech Recognition, Multi-Stream, Multiple Microphone Array, Two-Stage Training
\end{keywords}

\section{introduction}
\label{sec:intro}

The multi-stream paradigm in speech processing considers scenarios where parallel streams carry diverse or complementary task-related knowledge. 
In these cases, an appropriate strategy to fuse streams or select the most informative source is necessary. 
One potential source of inspiration in this setting is from the observations of parallel processing in the human auditory system, and resulting innovations have been successfully applied to conventional automatic speech recognition (ASR) frameworks \cite{mallidi2018practical, hermansky2013multistream, mallidi2016novel, hermansky2018coding}. 
For instance, multi-band acoustic modeling was formulated to address noise robustness \cite{mallidi2018practical, mallidi2016novel}. 
\cite{meyer2016performance} investigated several performance measures in spatial acoustic scenes to choose the most reliable source for hearing aids. The multi-modal applications combine visual \cite{palaskar2018end} or symbolic \cite{renduchintala2018multi} inputs together with audio signal to improve speech recognition.

The work that follows considers far-field ASR using multiple microphone arrays, a specific case of multi-stream paradigm. Without any knowledge of speaker-array distance or orientation, it is still challenging to speculate which array is most informative or least corrupted. 
The common methods of utilizing multiple arrays in conventional ASR are posterior combination \cite{wang2018stream, xiong2018channel}, ROVER \cite{fiscus1997post}, distributed beamformer \cite{yoshioka2019meeting}, and selection based on Signal-to-Noise/Interference Ratio (SNR/SIR) \cite{du2018theustc}. 

In recent years, with the increasing use of Deep Neural Networks (DNNs) in ASR, End-to-End (E2E) speech recognition approaches, which directly transcribe human speech into text, have received greater attention.  
The E2E models combine several disjoint components (acoustic model, pronunciation model, language model) from hybrid ASR into one single DNN for joint training.
Three dominant E2E architectures for ASR are Connentionist Temporal Classification (CTC) \cite{graves2006connectionist,graves2014towards,miao2015eesen}, attention-based encoder decoder \cite{chan2015listen,chorowski2015attention}, and Recurrent Neural Network Transducer (RNN-T) \cite{graves2012sequence,graves2013speech}. 
Coupled with a CTC network within a multi-task scheme, the joint CTC/Attention model \cite{kim2016joint_icassp2017,hori2017advances,watanabe2017hybrid} outperforms the attention-based model by addressing misalignment issues, achieving the state-of-the-art E2E performance on several benchmark datasets \cite{watanabe2017hybrid}.

In \cite{li2019multistream}, we proposed a novel multi-stream model based on a joint CTC/Attention E2E scheme, where each stream is characterized by a separate encoder and CTC network. 
A Hierarchical Attention Network (HAN) \cite{li2018multiencoder, wang2019stream} acts as a fusion component to dynamically assign higher weights for streams carrying more discriminative information for prediction.
The Multi-Encoder Multi-Array (MEM-Array) framework was introduced in \cite{li2019multistream} to improve the robustness of distant microphone arrays, where each array is represented by a separate encoder. 
While substantial improvements were reported within a two-stream configuration, there are two concerns when more streams are involved. First, during training, fitting all parallel encoders in device computing memory is potentially impractical for joint optimization, as the encoder is typically the largest component by far, i.e., 88\% of total parameters in this work.
Second, due to the data-hungry nature of DNNs and the expensive cost of collecting parallel data, training multiple models with excess degrees of freedom is not optimal.

In this paper, we present a practical two-stage training strategy on the MEM-Array framework targeting the aforementioned issues.
The proposed technique has the following highlights:
\begin{enumerate}
    \item In Stage-1, a single-stream model is trained using all data for better model generalization. The encoder will then acts as a Universal Feature Extractor (UFE) to process parallel data individually to generate a set of high-level parallel features. 
    \item Initializing components (CTC, decoder, frame-level attention) from Stage-1, Stage-2 training only optimizes the HAN component operating directly on UFE parallel features. The resulting memory and computation savings greatly simplify training, potentially allowing for more hyperparameter exploration or consideration of more complicated architectures. 
    \item Lack of adequate volume of data, specially parallel data, leads to overfit or is hard to tackle unseen data. The proposed two-stage strategy better defines the data augmentation scheme. Augmentation in Stage-1 aims to extract more discriminative high-level features and provides well-pretrained modules for Stage-2, whereas Stage-2 could focus on improving the robustness of information fusion.  
\end{enumerate}


\section{MEM-Array End-to-End Speech Recognition}
\label{sec:memarr}
In this session, we review the joint CTC/Attention framework and the extended MEM-Array model, one realization of the multi-stream approach with focus on distant multi-array scenario.

\subsection{Joint CTC/Attention}
\label{ssec:ctc/att}

The joint CTC/Attention architecture, illustrated in Stage-1 of Fig. \ref{fig:2stage}, takes advantage of both CTC and attention-based models within a Multi-Task Learning (MTL) scheme. 
The model directly maps a $T$-length sequence of $D$-dimensional speech vectors, $X=\{\textbf{x}_{t}\in \mathbb{R}^{D}|t = 1,2,...,T\}$, into an $L$-length label sequence, $C=\{c_{l}\in \mathcal{U}|l = 1,2,...,L\}$. 
Here $\mathcal{U}$ is a set of distinct labels. 
The encoder transforms the acoustic sequence $X$ into a higher-level feature representation $H=\{\textbf{h}_{1},..., \textbf{h}_{\lfloor T/s\rfloor}\}$, which is shared for the use of CTC and attention-based models.
Here, $\lfloor T/s\rfloor$ time instances are generated at the encoder-output level with a subsampling factor of $s$.
The loss function to be optimized is a logarithmic linear combination of CTC and attention objectives, i.e., $p_\textrm{ctc}(C|X)$ and $p_\textrm{att}^{\dagger}(C|X)$:
\begin{equation} 
\label{f:mtl}
\mathcal{L}_\textrm{MTL}=\lambda\log p_\textrm{ctc}(C|X)+(1-\lambda)\log p_\textrm{att}^{\dagger}(C|X),
\end{equation}
where $\lambda \in [0,1]$ is a hyper parameter. Note that $p_\textrm{att}^{\dagger}(C|X)$ is an approximated letter-wise objective where the probability of a prediction is conditioned on previous true labels. During inference, a label-synchronous beam search is employed to predict the most probable label sequence $\hat{C}$:
\begin{align}
\label{f:jointdec}
    \hat{C}=\arg\max_{C\in \mathcal{U}^{*}} &\{\lambda \log p_\textrm{ctc}(C|X)+(1-\lambda)\log p_\textrm{att}(C|X) \nonumber \\
    &+\gamma \log p_\textrm{lm}(C)\}, 
\end{align}
where $\log p_\textrm{lm}(C)$ is evaluated from an external Recurrent Neural Network Language Model (RNN-LM) with a scaling factor $\gamma$. 

\subsection{MEM-Array Model}
\label{ssec:memarr}

An end-to-end ASR model addressing the general multi-stream setting was introduced in \cite{li2019multistream}.
As one representative framework, MEM-Array concentrates on cases of far-field microphone arrays to handle different dynamics of streams. 
The architecture of $N$ streams is shown in Stage-2 of Fig. \ref{fig:2stage}.
Each encoder operates separately on a parallel input $X^{(i)}$ to extract a set of frame-wise hidden vectors $H^{(i)}$:
\begin{equation}
\label{eq:enc}
H^{(i)}=\textrm{Encoder}^{(i)}(X^{(i)}), i\in\{1, ..., N\},
\end{equation}
where we denote superscript $i$ as the index for stream $i$, and $H^{(i)}=\{\textbf{h}^{(i)}_{1},..., \textbf{h}^{(i)}_{\lfloor T^{(i)}/s\rfloor}\}$.
A frame-level attention mechanism is designated to each encoder to carry out the stream-specific speech-label alignment. 
For stream $i$, the letter-wise context vector $\textbf{r}_{l}^{(i)}$ is computed via a location-based attention network \cite{NIPS2015_5847} as follows:
\begin{equation}
\label{f:cv1}
\textbf{r}_{l}^{(i)}={\sum}_{t=1}^{\lfloor T^{(i)}/s^{(i)}\rfloor}a_{lt}^{(i)}\textbf{h}_{t}^{(i)},
\end{equation}
\begin{equation}
\label{f:cv2}
a^{(i)}_{lt} = \textrm{Attention}(\{a_{l-1}^{(i)}\}^{T^{(i)}}_{t=1}, \textbf{q}_{l-1}, \textbf{h}^{(i)}_t),  
\end{equation}
where ${a}^{(i)}_{lt}$ is the attention weight, a soft-alignment of $\textbf{h}^{(i)}_t$ for output $c_{l}$, and $\textbf{q}_{l-1}$ is the previous decoder state. 
In the multi-stream setting, the contribution of each stream changes dynamically. 
Hence, a secondary stream attention, the HAN component, is exploited for the purpose of robustness. 
The fusion context vector $\textbf{r}_l$ is obtained as a weighted summation of $\{\textbf{r}_l^{(i)}\}^{N}_{i=1}$:
\begin{equation}
\label{f:han}
\textbf{r}_{l}={\sum}_{i=1}^{N}\beta_{l}^{(i)}\textbf{r}_{l}^{(i)},
\end{equation}
\begin{equation}
\label{f:l2att}
\beta_{l}^{(i)}=\textrm{HierarchicalAttention}(\textbf{q}_{l-1}, \textbf{r}_l^{(i)}), i\in\{1, ..., N\}.
\end{equation}
where in this work, a content-based attention network \cite{NIPS2015_5847} is applied here, and $\beta_{l}^{(i)}$ is a Softmax output across $\{i\}^{N}_{1}$ from the HAN component, a stream-level attention weight for array $i$ of prediction $c_{l}$.
In addition, a separate CTC network is active for each encoder to enhance the stream diversity instead of sharing a CTC across all streams. 
In this setting, the MEM-Array model follows Eq. (\ref{f:mtl}) and (\ref{f:jointdec}) with a modified CTC objective:
\begin{equation}
\log p_\textrm{ctc}(C|X)=\frac{1}{N}{\sum}_{i=1}^{N}\log p_{\textrm{ctc}^{(i)}}(C|X),
\end{equation}
where joint CTC loss is the average of per-encoder CTCs. 

\section{Proposed Training Strategy}

\begin{figure}[htp]
\vspace{-0.27cm}
  \centering 
  \centerline{\includegraphics[width=9.1cm]{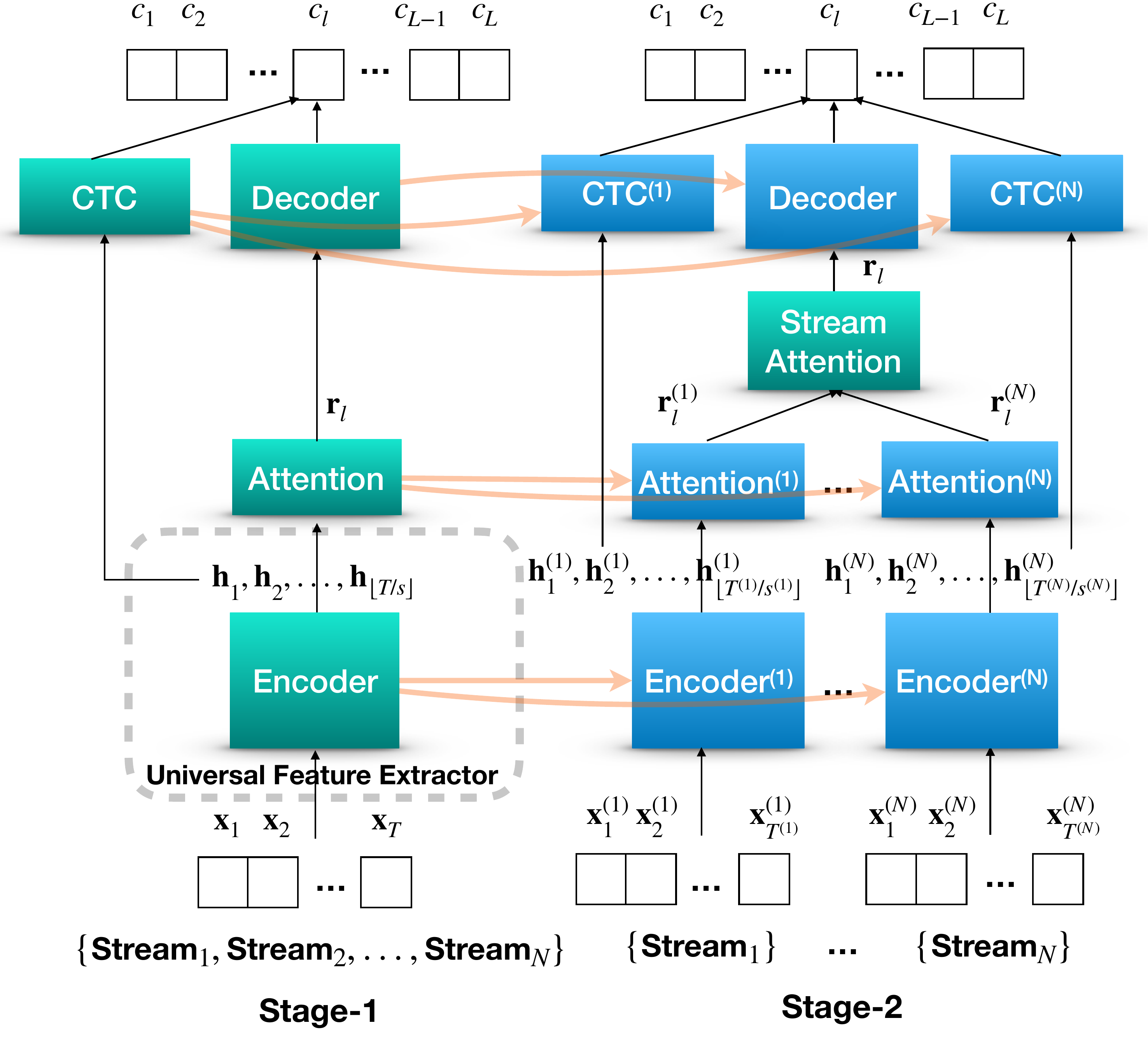}}
  \caption{Proposed Two-Stage Training Strategy. Color ``green'' indicates the components are trainable; Color ``blue'' means parameters of the components are frozen. }
  \label{fig:2stage}
  \vspace{-0.1cm}

\end{figure}

In this section, we present a practical two-stage training strategy for the MEM-Array model, depicted in Fig. \ref{fig:2stage}. The details of each stage are discussed in the following sections.

\label{sec:twostage}

\subsection{Stage 1: Universal Feature Extractor}
\label{ssec:stage1}

The intent of Stage-1 is to obtain a single well-trained encoder, which we refer to as Universal Feature Extractor (UFE), to prepare a new set of high-level features for Stage-2. 
Encoder in E2E model can be viewed as an acoustic modeling that generates sequences $H=\{\textbf{h}_{1},..., \textbf{h}_{\lfloor T/s\rfloor}\}$ with more discrinimative power for prediction. 
We denote the encoder outputs $H$ as the UFE features. 
In general, the majority of the overall parameters are contained in the encoder. 

In Stage-1, a single-stream joint CTC/Attention model is optimized as shown in Fig. \ref{fig:2stage}.
Audio features from all available streams are used to train the model. 
After training, we extract UFE features $H^{(i)}=\{\textbf{h}^{(i)}_{1},..., \textbf{h}^{(i)}_{\lfloor T^{(i)}/s\rfloor}\}$ for each stream $i$, separately. 
Since subsampling mitigates the increased dimension of UFE features, it is possible to save the UFE features at a similar size to the original speech features. 
Moreover, byproducts in Stage-1, such as decoder, CTC and attention, can be used for initialization in Stage-2. 

\subsection{Stage 2: Parallel-Stream Fusion}
\label{ssec:stage2}

As illustrated in Fig. \ref{fig:2stage}, Stage-2 focuses on training the fusion component within the multi-stream context.
The MEM-Array model uses parallel encoders as information streams. 
The previous strategy uses joint training with multiple large encoders, which is expensive in memory and time for more complex models or more streams. 
Taking advantage of UFE features greatly alleviates this complication.

In Stage-2, we formulate a multi-stream scheme on UFE features $\{H^{(i)}\}^N_{i=1}$ as parallel inputs.
In this model, parameters of all components, except the stream attention module, are initialized from Stage-1 and frozen during optimization. 
The stream fusion component is randomly initialized, and is the only trainable element in Stage-2. 
Without any involvement of encoders, frame-level attention directly operates on UFE features. 
This setup not only reduces the amount of required parallel data, but it also greatly reduces memory and time requirements, allowing for more thorough hyperparameter exploration or utilization of more complex architectures.

\section{Experimental Setup}
\label{sec:exptsetup}

We demonstrated the two-stage training strategy using two datasets: DIHRA English WSJ \cite{ravanelli2016realistic} and AMI Meeting Corpus \cite{carletta2005ami}. 

The DIRHA English WSJ is part of the DIRHA project, which focuses on speech interaction in domestic scenes via distant microphones. 
There are in total 32 microphones placed in an apartment setting with a living room and a kitchen. We chose a 6-mic circular array (Beam Circular Array) and an 11-mic linear array (Beam Linear Array) in the living room for experiments with two parallel streams. 
Additionally, a single microphone (L1C) was picked to serve as a third stream in 3-stream experiments. 
Training data was created by contaminating original Wall Street Journal data (WSJ0 and WSJ1, 81 hours per stream), providing room impulse responses for each stream. 
Simulated WSJ recordings with domestic  background  noise  and  reverberation were used as the development  set for cross  validation. 
The evaluation set has 409 WSJ recordings uttered in real domestic scenario. 

The AMI Meeting Corpus was collected in three instrumented rooms with meeting conversations. 
Each room has two microphone arrays to collect 100 hours of far-field signal-synchronized recordings.
With no speakers overlapping, the training, development and evaluation set have 81 hours, 9 hours and 9 hours of meeting recordings, respectively. 
No close-talk microphone recordings are used here.

We designed both 2-stream and 3-stream settings for DIRHA and 2-stream experiments for AMI. Note that for each array, the multi-channel input was synthesized into single-channel audio using Delay-and-Sum beamforming with BeamformIt \cite{anguera2007acoustic}.
Experiments were conducted using a Pytorch back-end on ESPnet \cite{watanabe2018espnet} configured as described in Table \ref{tab:config}.

\begin{table}[th]
  \begin{center}
   	\caption{Experimental Configuration.}
    \label{tab:config}
	\begin{tabular}{ll}
	  \toprule
	  \toprule
	  {\scriptsize{}{\bf Feature}} & {\scriptsize{}80-dim log-mel filter bank + 3-dim pitch}\\
      \hline
	  {\scriptsize{}{\bf Model}} \\
      {\scriptsize{}Encoder type} & {\scriptsize{}VGGBLSTM \cite{hori2017advances, cho2018multilingual} (subsampling factor: 4)}\\
      {\scriptsize{}Encoder layers} & {\scriptsize{}6(CNN)+2(BLSTM)}\\
      {\scriptsize{}Encoder units } & {\scriptsize{}320 cells (BLSTM layers)}\\
      {\scriptsize{}Encoder projection} & {\scriptsize{}320 cells (BLSTM layers)}\\
      {\scriptsize{}Frame-level Attention} & {\scriptsize{}320-cell Content-based}\\
      {\scriptsize{}Stream Attention} & {\scriptsize{}320-cell Location-based}\\
      {\scriptsize{}Decoder type} & {\scriptsize{}1-layer 300-cell LSTM}\\
      \hline 
	  {\scriptsize{}{\bf Train and Decode}} \\
	 {\scriptsize{}Optimizer} & {\scriptsize{}AdaDelta (Batch size: 15)}\\
	 {\scriptsize{}Training Epoch} & {\scriptsize{}30 epochs (patience:3 epochs)}\\
      {\scriptsize{}CTC weight $\lambda$} & {\scriptsize{}0.2 (train); 0.3
      (decode)}\\
    {\scriptsize{}Label Smoothing} & {\scriptsize{}Type: Unigram \cite{pereyra2017regularizing}, Weight: 0.05}\\
        \hline 
      {\scriptsize{}{\bf RNN-LM}} \\
      {\scriptsize{}Type} & {\scriptsize{}Look-ahead Word-level RNNLM \cite{hori2018end}}\\
      {\scriptsize{}Train data} & {\scriptsize{}AMI:AMI; DIRHA:WSJ0-1+extra WSJ text}\\
      {\scriptsize{}LM weight $\gamma$} & {\scriptsize{}AMI:0.5; DIRHA:1.0}\\
      \bottomrule 
      \bottomrule
	\end{tabular}
  \end{center}
    \vspace{-0.65cm}
\end{table}

\section{results and discussions}
\label{sec:expts}

Firstly, we examined UFE features in a single-stream setting.  Next, the full proposed strategy was analyzed in comparison to the previous approach as well as to several conventional fusion methods on DIRHA 2-stream case. 
Results on AMI and extension with more streams on DIRHA were explored as well. 
Lastly, we considered the potential benefits of data augmentation in this framework.  

\subsection{Effectiveness of Two-Stage Training}

In this section, we discuss the results on 2-stream DIRHA to demonstrate the value of proposed strategy. 
First, to evaluate Stage-1 training, Character/Word Error Rates (CER/WER) results on single stream systems are summarized in Table \ref{tab:single-stream}. 
Training the model using data from both streams improves performance substantially on the individual arrays, i.e., $37.6\%\rightarrow{}33.9\%$ and $39.2\%\rightarrow{}30.7\%$. 
The UFE features are the outputs of an encoder trained with this improved strategy.
In our setup, 320-dimensional UFE features took slightly smaller space than 83-dimensional acoustic frames since the subsampling factor $s=4$. 

\begin{table}[!htbp]
  \begin{center}
   	\caption{Stage-1 results on 2-stream DIRHA.}
	\begin{tabular}{lcccc}
	  \toprule
	  \toprule
	   &  \multicolumn{2}{c}{$\text{Arr}_1$} & \multicolumn{2}{c}{$\text{Arr}_2$}\\
	  Train Data & CER(\%) & WER(\%) & CER(\%) & WER(\%)  \\
      \midrule
      \multicolumn{3}{l}{{\it Single Stream}} & \\
      $\text{Arr}_1$ & 22.3& 37.6 & -- & -- \\
      $\text{Arr}_2$ & --&--&23.0&39.2 \\
      $\text{Arr}_1, \text{Arr}_2$ & \textbf{20.1} &\textbf{33.9}&\textbf{17.9}&\textbf{30.7} \\
      \bottomrule
      \bottomrule
	\end{tabular}
	\label{tab:single-stream}
  \end{center}
    \vspace{-0.6cm}
\end{table}

Table \ref{tab:two-stage} illustrates several training strategies in Stage-2. 
Since Stage-2 operates on UFE features directly, its training only involves, at most, frame-level attention (ATT), decoder (DEC), hierarchical attention (HAN) and CTC. 
These experiments considered which of these components should be initialized from their Stage-1 counterparts, as well as which components should be fine-tuned or frozen during Stage-2 updates.
In both cases of fine-tuning or freezing Pre-Trained (PT) modules in Stage-2, more noticeable improvements were reported with introducing more pretraining knowledge, i.e., $32.9\%\rightarrow{}28.4\%$ and $31.8\%\rightarrow{}26.8\%$, respectively.
Moreover, keeping all PT components frozen during Stage-2 and training solely the fusion module showed relative WER reduction of 5.6\% ($28.4\%\rightarrow{}26.8\%$) with only 0.2 million active parameters. 
Overall, a substantial improvement of 18.8\% relative WER reduction ($33.0\%\rightarrow{}26.8\%$) was observed compared to jointly training a massive model, including encoders, from scratch. 

\begin{table}[!htbp]
  \begin{center}
   	\caption{WER(\%) Comparison among various Stage-2 training strategies on 2-stream DIRHA. Note that components with random initialization in Stage-2 are listed in parentheses of first column. The amount of trainable parameters in Stage-2 when freezing Stage-1 Pre-Trained (PT) components is stated in parentheses of last column.} 
	\begin{tabular}{lcc}
	  \toprule
	  \toprule
     Initialization with & Fine-tune & Freeze \\
	 PT Comp. \scriptsize{(rand. init. comp.)}  & PT Comp.&  PT Comp.  \\
      \midrule
          {\it No Two-Stage} \\
      Baseline  & -- &33.0 (21.82M) \\
\midrule
      {\it Two-Stage} \\
      -- \scriptsize{(ATT, DEC, CTC, HAN)} & 32.9&31.8 (1.78M) \\
      CTC \scriptsize{(ATT, DEC, HAN)} &34.4&30.7 (1.75M) \\
      ATT \scriptsize{(DEC, CTC, HAN)} & 33.3& 30.6 (1.37M)\\
      ATT, DEC \scriptsize{(CTC, HAN)} &29.0 &27.4 (0.23M) \\
      ATT, DEC, CTC \scriptsize{(HAN)} & \textbf{28.4}&\textbf{26.8} (0.20M) \\

      \bottomrule
      \bottomrule
	\end{tabular}
	\label{tab:two-stage}
  \end{center}
    \vspace{-0.8cm}
\end{table}

\subsection{Multi-Stream v.s. Conventional Methods}

In Table \ref{tab:more}, the MEM-Array model with our two-stage training strategy consistently outperforms the baseline model which needs joint training after random initialization. 
18.8\%, 32.4\%, and 8.2\% relative WER reductions are achieved in 2-stream DIRHA, 3-stream DIRHA, and 2-stream AMI, respectively.
Note that AMI experients were conducted using VGGBLSTM with 2-layer BLSTM layers without any close-talk recordings and data perturbations. 
It is worth mentioning that those reductions in WERs were accomplished while simultaneously significantly decreasing the number of unique parameters in training by avoiding costly multiples of the large encoder component (10 million parameters per stream, in this case).

In addition, results from several conventional fusion strategies are shown in Table \ref{tab:more}: signal-level fusion via WAV alignment and average; feature-level frame-by-frame concatenation; word-level prediction fusion using ROVER. 
For fair comparison, single-level and word-level fusion models utilized Stage-1 pre-trained models as their initialization. 
Note that word-level fusion operates on decoding results from pretrained single-stream from Stage-1. 
Still, our proposed strategy consistently perform better than all other fusion methods in all conditions.



\begin{table}[!htbp]
  \begin{center}
   	\caption{WER(\%) Comparson between proposed two-stage approach and alternative conventional methods.}
	\begin{tabular}{lcccc}
	  \toprule
	  \toprule
         & Unique & \multicolumn{3}{c}{\# Streams} \\
	  	 & Params &  \multicolumn{2}{c}{DIRHA}&AMI \\
	 Model &  (in million)  &  2   & 3 &2\\
      \midrule
      \multicolumn{3}{l}{{\it MEM-Array Model}}\\
      Baseline \cite{li2019multistream} &21.8(2),32.1(3) &33.0& 32.1& 59.5 \\
      Proposed Strategy &11.6& \textbf{26.8}&\textbf{21.7}& \textbf{54.6}\\
      \midrule
      \multicolumn{3}{l}{{\it Other Fusion Methods}}\\
        WAV Align.\& Avg. &11.4 & 32.4& 30.1& 55.9\\
      Frame Concat.& 16.9(2),23.8(3) &33.7 &33.8 &59.4\\
          ROVER & 11.4 & 34.2& 23.6&58.0\\
      \bottomrule
      \bottomrule
	\end{tabular}
	\label{tab:more}
  \end{center}
  \vspace{-0.6cm}
\end{table}

\subsection{Discussion on Data Augmentation}

The two-stage training strategy provides various opportunities for data augmentation. 
Stage-1 does not consider parallel data, so any augmentation technique for regular E2E ASR could be applied in this stage to improve the robustness of the UFE.
Stage-2 augmentation, on the other hand, would be expected to improve robustness of the combination of corrupted individual streams.
In this study, we employed a simple data augmentation technique called SpecAugment \cite{Park2019SpecAugmentAS}, which randomly removes sections of the input signal in a structured fashion, to demonstrate the potential of this direction. 
Table \ref{tab:specaug} shows the improvements from applying SpecAugment on two separate training stages. 
The best performance was from data augmentation on Stage-1 when freezing all Stage-1 pretrained components. With additional Stage-2 SpecAugment, there was not a noticeable difference in terms of WERs ($22.6\%$ v.s. $22.4\%$ and $22.6\%$ v.s. $22.5\%$). 
10\% absolute WER reduction was achived in AMI with two stage augmentation.
However, it is important to remember that, while the performance gap from fine-tuning versus freezing pre-trained components is narrowed with Stage-2 augmentation, the reductions in Stage-2 memory and computation requirements are still substantially better with frozen parameters.

\begin{table}[!htbp]
  \begin{center}
  	\caption{Perfermance (WER(\%)) investigation of two-stage data augmentation using SpecAugment on 2-stream DIRHA and AMI.}
	\begin{tabular}{lccc}
	  \toprule
	  \toprule
	  	  	    &\multicolumn{2}{c}{DIHRA} & \\
	  	    &Fine-tune& Freeze & \\

	 Model   &PT Comp.& PT Comp. & AMI\\
      \midrule
      {\it Augmentation}& & &\\
      no SpecAugment &  28.4 & 26.8&59.5 \\
      Stage-1 &22.6& \textbf{22.4}&55.8\\
      Stage-1, Stage-2 &22.5&22.6&\textbf{49.2}\\
      \bottomrule
      \bottomrule
	\end{tabular}
	\label{tab:specaug}
  \end{center}
    \vspace{-0.8cm}
\end{table}




\section{conclusions}
\label{sec:concl}

In this work, we proposed a practical two-stage training strategy to improve multi-stream end-to-end ASR. 
A universal feature extractor is trained in Stage-1 with all available data. 
In Stage-2, a set of high-level UFE features are used to train a multi-stream model without requiring highly-parameterized parallel encoders.
This two-stage strategy remarkably alleviates the burden of optimizing a massive multi-encoder model while still substantially improving the ASR performance.
This work shows great potential and value for this approach, but numerous directions remain for future exploration.
More sophisticated data augmentation techniques beyond the single method considered here should be explored.
Stage-2 training could also possibly benefit from stream-specific knowledge by exploiting more complex stream attention.
Strategies for adding new streams to an existing model would also be worth investigating.

\bibliographystyle{IEEEbib}
\bibliography{strings,refs}

\end{document}